\documentclass{article}

    \PassOptionsToPackage{numbers, compress}{natbib}

\usepackage[preprint]{neurips_2025}

\usepackage[utf8]{inputenc} 
\usepackage[T1]{fontenc}    
\usepackage{hyperref}       
\usepackage{url}            
\usepackage{booktabs}       
\usepackage{amsfonts}       
\usepackage{nicefrac}       
\usepackage{microtype}      
\usepackage{xcolor}         

\usepackage{multirow}
\usepackage{makecell}
\usepackage{pifont}
\usepackage{algorithm}
\usepackage{algorithmic}
\usepackage{amsmath}
\usepackage{amssymb}

\usepackage{graphicx}

\usepackage{wrapfig}

\title{UniForward: Unified 3D Scene and Semantic Field Reconstruction via Feed-Forward Gaussian Splatting from Only Sparse-View Images}

\author{%
  Qijian Tian \\
  Shanghai Jiao Tong University \\
  Shanghai, China \\
  \texttt{tianqijian@sjtu.edu.cn} \\
  \And
  Xin Tan \\
  East China Normal University \\
  Shanghai, China \\
  \texttt{xtan@cs.ecnu.edu.cn} \\
  \And
  Jingyu Gong \\
  East China Normal University \\
  Shanghai, China \\
  \texttt{jygong@cs.ecnu.edu.cn} \\
  \And
  Yuan Xie \\
  East China Normal University \\
  Shanghai, China \\
  \texttt{xieyuan8589@foxmail.com} \\
  \And
  Lizhuang Ma \\
  Shanghai Jiao Tong University \\
  Shanghai, China \\
  \texttt{ma-lz@cs.sjtu.edu.cn} \\
}

\begin{document}

\maketitle

\begin{abstract}
We propose a feed-forward Gaussian Splatting model that unifies 3D scene and semantic field reconstruction. Combining 3D scenes with semantic fields facilitates the perception and understanding of the surrounding environment. However, key challenges include embedding semantics into 3D representations, achieving generalizable real-time reconstruction, and ensuring practical applicability by using only images as input without camera parameters or ground truth depth. To this end, we propose UniForward, a feed-forward model to predict 3D Gaussians with anisotropic semantic features from only uncalibrated and unposed sparse-view images. To enable the unified representation of the 3D scene and semantic field, we embed semantic features into 3D Gaussians and predict them through a dual-branch decoupled decoder. During training, we propose a loss-guided view sampler to sample views from easy to hard, eliminating the need for ground truth depth or masks required by previous methods and stabilizing the training process. The whole model can be trained end-to-end using a photometric loss and a distillation loss that leverages semantic features from a pre-trained 2D semantic model. At the inference stage, our UniForward can reconstruct 3D scenes and the corresponding semantic fields in real time from only sparse-view images. The reconstructed 3D scenes achieve high-quality rendering, and the reconstructed 3D semantic field enables the rendering of view-consistent semantic features from arbitrary views, which can be further decoded into dense segmentation masks in an open-vocabulary manner. Experiments on novel view synthesis and novel view segmentation demonstrate that our method achieves state-of-the-art performances for unifying 3D scene and semantic field reconstruction.\par

\end{abstract}

\section{Introduction}
3D scene reconstruction is a critical task in computer vision. Advances in 3D reconstruction enhance realism for embodied AI and robotics and enable agents to synthesize novel views from observations, improving perception and understanding of the surrounding environment~\cite {embodied_ai, embodied_ai_2}.\par

Neural Radiance Fields (NeRF)~\cite{nerf} and 3D Gaussian Splatting (3DGS)~\cite{3dgs} have recently made significant progress in 3D reconstruction. Compared to NeRF-based methods, 3DGS-based methods are more widely adopted for their efficiency and expressive representation.  
However, these methods remain offline as they require pre-scene optimization, limiting their generalizable capability and failing to meet the real-time requirements of downstream tasks, such as robot navigation~\cite{robotics}. \par

Subsequently, several feed-forward Gaussian Splatting methods~\cite{pixelsplat, mvsplat, splatt3r, noposplat, drivinggaussian} have been proposed to explore generalizable 3D reconstruction from sparse-view images. These methods leverage prior knowledge learned from large-scale datasets and require only a feed-forward pass during inference, enabling generalizable real-time reconstruction without per-scene optimization. \par


\begin{figure}[t!]
    \centering
    \includegraphics[width=0.8\linewidth]{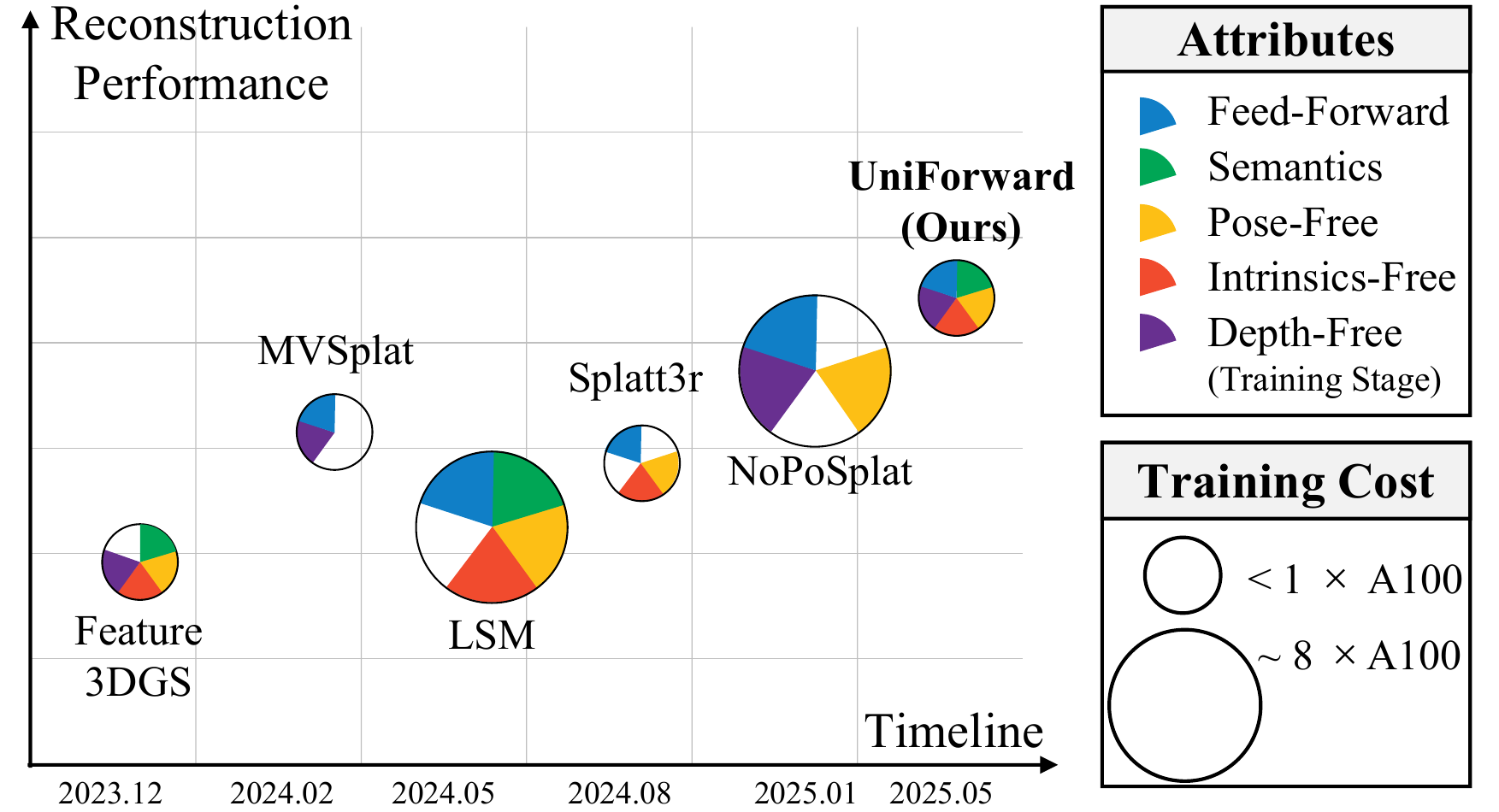}
    \caption{Comparison of our UniForward with the latest related works. Our UniForward unifies 3D scene and semantic field reconstruction with fewer constraints.}
    \label{fig:teaser}
\end{figure}

Although feed-forward methods achieve generalizable real-time reconstruction, they recover only geometry and appearance, lacking semantics. However, semantics also play a critical role in decision-making. For example, an agent needs to perceive its environment from limited, sparse views and locate concepts such as ``door''. This process requires embedding semantics directly into the 3D scene through a unified framework that jointly reconstructs both the 3D scene and its semantic field, enabling continuous and view-consistent semantics to support coherent spatial reasoning. While existing semantic models extract semantics only from observed 2D views, they fail to embed semantics into 3D representation, limiting novel view inference and preventing comprehensive 3D scene understanding. Previous NeRF-based methods~\cite{semanticnerf, nvos, panopticligting, dff, nerfsos, lerf, n3f, semanticray} have explored embedding semantics into NeRF, but they are inefficient due to implicit representation and struggle to handle sparse view inputs. Recent advances in 3DGS~\cite{ legaussians, langsplat, feature3dgs, fastlgs} embed semantics into 3DGS, while these methods only focus on per-scene optimization and fail to generalize across scenes. A recent work, LSM~\cite{lsm}, attempts to estimate geometry, appearance, and semantics in a single feed-forward pass. However, it yields suboptimal performance in both scene and semantic reconstruction due to direct regression, and the reliance on depth for training limits scalability to large-scale video applications.\par

Our goal is to unify 3D scene and semantic field reconstruction, forming a unified 3D representation of geometry, appearance, and semantics. To ensure practicality and support downstream embodied AI and robotics applications, this task faces several challenges: 1) embedding semantics into 3D representation; 2) achieving generalizable, real-time reconstruction; 3) taking only images as input without camera parameters (camera pose and intrinsics) or ground truth depth. \par

To this end, we propose \textbf{UniForward}, a \textbf{feed-forward} Gaussian Splatting model that \textbf{unifies 3D scene and semantic field reconstruction from only sparse-view images}. As illustrated in Figure~\ref{fig:teaser}, it achieves the feed-forward reconstruction of 3D scenes and semantics using only raw images, without camera parameters or ground truth depth, while existing methods~\cite{feature3dgs, mvsplat, lsm, splatt3r, noposplat} struggle to address all of these challenges. Following recent advances in 3D point cloud reconstruction~\cite{dust3r, mast3r}, we adopt the paradigm that directly regresses points from different views in the same coordinates, achieving generalizable pose-free reconstruction in real time. However, existing methods~\cite{dust3r, mast3r, splatt3r, noposplat} only predict 3D points or Gaussians without incorporating semantic information. Embedding semantics into 3D representations is non-trivial, and simply regressing geometry, appearance, and semantics together only leads to suboptimal performance as in the previous method~\cite{lsm}. To this end, we propose a \textbf{dual-branch decoupled decoder} that separates the prediction of geometry, appearance, and semantic attributes. This design is motivated by the observation that the positions of Gaussians and the camera poses primarily determine the geometry of the reconstructed scene, while the remaining Gaussian attributes and semantic features relate to appearance and semantics. Accordingly, we construct a geometry branch to predict scene geometry and an attribute branch to model appearance and semantic information. Each branch further includes separate heads for predicting geometry, appearance, and semantic attributes. Besides, some previous pose-free feed-forward reconstruction methods~\cite{splatt3r, lsm} rely on ground truth depth or masks for training, which limits the usage of large-scale video data. However, directly discarding depth ground truth or masks may cause unseen regions in novel views to affect model training, resulting in incorrect expansion of Gaussians at scene boundaries. To this end, we propose a \textbf{loss-guided view sampler} that helps the model adapt from large to small view overlap and stabilize training without requiring depth or masks. The whole model can be \textbf{optimized end-to-end} via novel-view rendering of images and semantic feature maps, using a photometric loss and a distillation loss that distills semantics from a pre-trained 2D semantic model~\cite{lseg}. At the inference stage, the trained model requires only sparse-view images to reconstruct a 3D scene and its semantic field in approximately 0.1s, while generalizing across diverse scenes. Utilizing the reconstructed scene, we can render view-consistent images and feature maps from novel views. The feature maps can be further decoded into the text-guided novel view segmentation masks. \par

We summarize our main contributions as follows:
\begin{itemize}
\item We propose UniForward, a feed-forward Gaussian Splatting model that unifies 3D scene and semantic field reconstruction from uncalibrated and unposed sparse views. A dual-branch decoupled decoder is proposed to predict Gaussians with semantic embedding. \par
\item We propose a loss-guided view sampler to automatically select views from easy to hard during training, which eliminates the need for ground truth depth and masks, and enhances training stability. \par
\item Our UniForward achieves state-of-the-art performance on both novel view synthesis and novel view segmentation, demonstrating its capability to unify 3D scene and semantic field reconstruction. \par
\end{itemize}

\section{Related Work}

\subsection{Feed-Forward Reconstruction}
The original reconstruction methods, such as NeRF~\cite{nerf} and 3DGS~\cite{3dgs}, are per-scene optimized since they require dozens of images for reconstructing a specific scene and lack generalizability across different scenes.
In contrast, feed-forward reconstruction methods~\cite{pixelnerf, pixelsplat, mvsplat, drivingforward} leverage powerful priors learned from large-scale datasets to achieve generalizable real-time reconstruction. DUSt3R~\cite{dust3r} and MASt3R~\cite{mast3r} further propose a novel paradigm to directly predict 3D point clouds from unposed image pairs, achieving pose-free feed-forward reconstruction. Built upon the paradigm, Splatt3R~\cite{splatt3r} and NoPoSplat~\cite{noposplat} extend the 3D representation to 3DGS. Our UniForward also achieves the pose-free feed-forward reconstruction, but we step further to embed the semantics into the 3D representation, enabling the comprehensive understanding of the surrounding environment. \par

\subsection{Embedding Semantics into 3D}
Simultaneously reconstructing 3D scenes and semantic fields has been explored using NeRF and 3DGS. Some NeRF-based methods embed semantics into NeRF using 2D semantic labels~\cite{semanticnerf, nvos, panopticligting}. Other methods attempt to lift the feature from 2D models to construct semantic fields instead of using semantic labels~\cite{dff, nerfsos, lerf, n3f, semanticray}. Although these methods successfully embed semantics into NeRF, the implicit representation limits their efficiency. Recent advances in 3DGS greatly improve the efficiency of reconstruction and rendering, and some methods leverage 3DGS to embed semantics into 3D scenes~\cite{langsplat, legaussians, feature3dgs, fastlgs}. Although 3DGS-based methods improve the efficiency of embedding semantics into 3D compared to NeRF-based methods, these methods require per-scene optimization and lack generalization capabilities, thus limiting their application in areas such as embodied AI. A recent method, LSM~\cite{lsm}, introduces a generalizable reconstruction model to estimate geometry, appearance, and semantics. However, it achieves suboptimal performance on both novel view synthesis and segmentation despite training with ground truth depth on 8 A100 GPUs for 3 days. In contrast, our UniForward eliminates the need for depth and achieves superior reconstruction quality for both 3D scenes and semantic fields with less training cost. \par

\section{Method}

\begin{figure}[ht]
    \centering
    \includegraphics[width=0.95\linewidth]{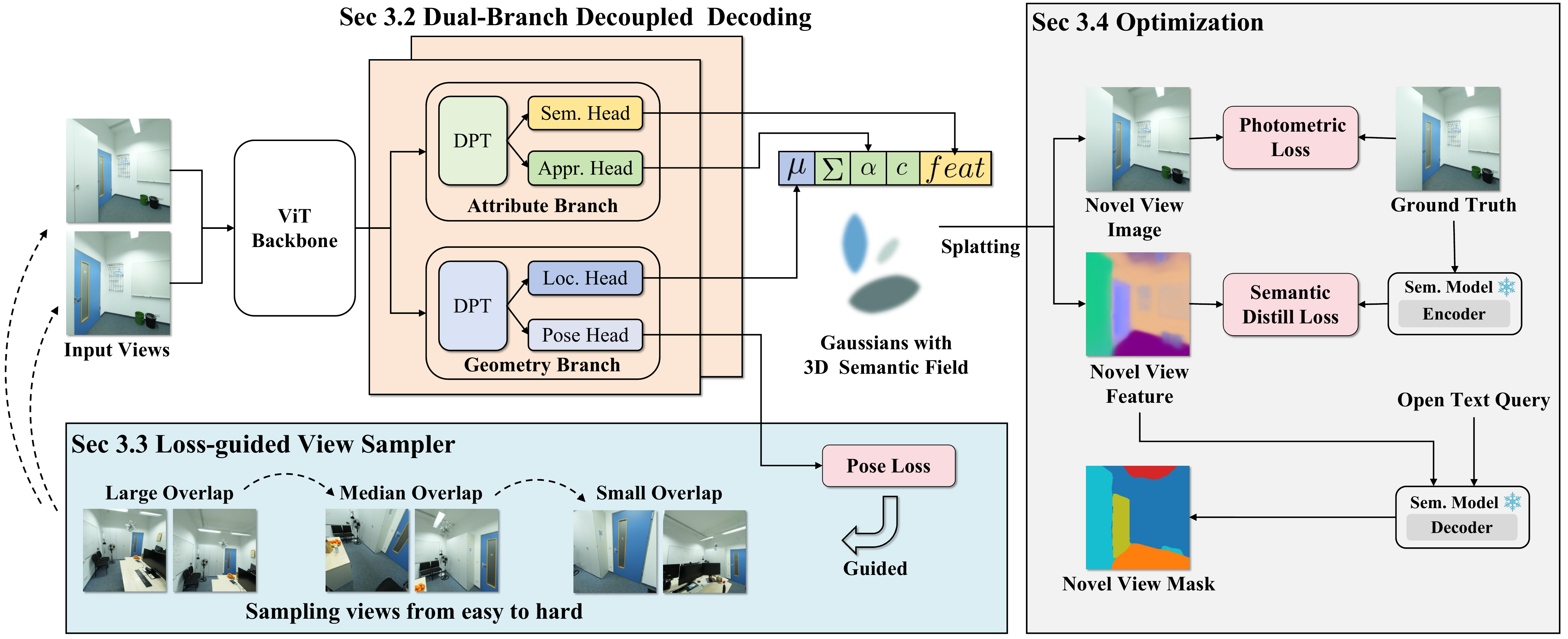}
    \caption{\textbf{Overview of UniForward}. Built upon the pose-free feed-forward reconstruction framework, the proposed dual-branch decoupled decoder separately predicts geometry, appearance, and semantic attributes of Gaussians, embedding semantics into the 3D representation. During training, a loss-guided view sampler selects training views from easy to hard. The model can be optimized through photometric loss and semantic distillation loss without ground truth depth. During inference, UniForward enables generalizable real-time 3D scenes and semantic fields reconstruction from only sparse-view images.}
    \label{fig:main}
\end{figure}

\subsection{Overview}

Our UniForward takes paired sparse-view images as input, aiming to unify 3D scene and semantic field reconstruction in a feed-forward manner. The overall framework is illustrated in Figure~\ref{fig:main}. Following recent advances in 3D point cloud reconstruction~\cite{dust3r, mast3r}, we use a ViT-based backbone to extract image features and predict Gaussians in a shared coordinate space from each view, enabling pose-free feed-forward reconstruction. To further embed the semantics into the 3D representation, we use 3D Gaussians with anisotropic semantic features. The components of 3D Gaussians (i.e. the position $\mu$, the 3D covariance matrix $\Sigma$, the opacity $\alpha$, color $c$ represented by spherical harmonics, and the additional semantic feature $feat$) are predicted by the proposed \textbf{dual-branch decoupled decoder (Sec.~\ref{dual})}, decoupling geometry, appearance, and semantics for more accurate, unified reconstruction of both the 3D scene and semantic field. We also propose a \textbf{loss-guided view sampler (Sec.~\ref{sampler})} to sample views from easy to hard, which eliminates the need for ground truth depth or masks during training and enables the usage of large-scale video data. The whole model can be \textbf{optimized end-to-end (Sec.~\ref{optimization})} with a photometric loss and a distillation loss that leverages semantic features from a pre-trained 2D semantics model. At the inference stage, our UniForward simultaneously reconstructs 3D scenes and semantic fields from uncalibrated and unposed sparse views in real time and can generalize across different scenes.\par

\subsection{Dual-branch Decoupled Decoder}\label{dual}

By extracting features of each view from their corresponding backbones, we aim to decode them to obtain Gaussians with semantic features in the same coordinates, thereby unifying the 3D scene and semantic field reconstruction. However, different Gaussian components correspond to different aspects of the scene: the position $\mu$ specifies geometry; the 3D covariance matrix $\Sigma$, opacity $\alpha$, and color $c$ specify appearance; and the additional semantic feature $feat$ specifies semantics. To accurately predict each component, we decouple them into two groups and design two branches: the geometry branch and the attribute branch. Each branch consists of a DPT-based decoder~\cite{dpt} and the following component-specific prediction heads. \par

\subsubsection{Geometry Branch}

The geometry branch focuses on recovering scene geometry by predicting Gaussian positions. Since we predict pixel-wise Gaussians, geometry is inherently tied to the input view poses. Accurately estimating positions across views implies that the model has learned the relative poses of input views. Therefore, we include two heads in the geometry branch: the localization head for Gaussian position prediction and the pose head for predicting relative poses as an auxiliary task to enhance geometry prediction, which can be denoted as:
\begin{equation}
\text{Geometry Branch} \left\{
\begin{aligned}
\text{Localization Head}(D(f^v)) & \rightarrow \{\mu_i^v\}_{i=1}^{h \times w} \\
\text{Pose Head}(D(f^v)) & \rightarrow [R | T]^{v \rightarrow 0} 
\end{aligned}
\right.
\quad
v \in \{0, 1\},
\end{equation}\par
where $D$ is the DPT decoder, $v$ is the index of view, $f$ is the feature extracted through backbone, $h$ and $w$ are the height and width of the image, $\mu_{i}$ is the position of Gaussian center, and $[R |T]$ is the relative pose to the coordinate of the first view. 

\textbf{Localization Head}. The localization head outputs the 3D position of each Gaussian’s center. Recent pose-required feed-forward methods~\cite{pixelsplat, mvsplat} first estimate depth for each view and then project depth into 3D space using known camera extrinsics. However, for pose-free reconstruction, the extrinsics are unknown, and using estimated extrinsics in this depth-then-projection pipeline may lead to error accumulation and degrade reconstruction quality. To this end, following recent 3D point cloud reconstruction methods~\cite{dust3r, mast3r}, we adopt the paradigm that directly predicts the pixel-wise Gaussians' center position from each view in the same coordinate. To be specific, each view is processed by its own dual-branch decoupled decoder. The first view’s localization head predicts the first view's pixel-wise Gaussian center position in its local camera coordinate system, which we refer to as the canonical space, while the second view’s localization head predicts the second view's pixel-wise Gaussian center position still in the same canonical space, i.e., the first view’s local camera coordinate system. Therefore, the Gaussians from different views exist in the same space, constituting the geometric structure of the 3D scene. \par

\textbf{Pose Head}. Since the canonical space is the first view's local camera coordinate, we introduce a pose head alongside the localization head in the geometry branch of the second view, which directly estimates the relative poses of input views. The motivation is to fully leverage extrinsic information and directly recover relative camera poses without any post-processing from the predicted Gaussians. By serving as an auxiliary proxy task for Gaussian position prediction, it enhances the model’s understanding of the geometric relationships between input views. During training, the pose head is directly supervised from ground truth extrinsics. The relative poses are represented by a rotation matrix and a translation vector. The rotation matrix is termed as a normalized quaternion since each rotation matrix corresponds to only one normalized quaternion (with real part being positive). We utilize the L2 loss with ground truth translation vector $x$ and quaternion $q$:
\begin{equation}
    \mathcal{L}_{pose} = \lVert \hat{x} - x \rVert_{2} + \lVert \hat{q} - \frac{q}{\lVert q \rVert} \rVert_{2},
\end{equation}
where $\hat{x}$ and $\hat{q}$ are the estimated translation vector and quaternion, which constitute the estimated relative camera pose. Note that we do not introduce additional supervision signals since extrinsics are also used in the following photometric loss. This follows the standard practice in existing pose-free feed-forward reconstruction methods~\cite{pixelsplat, mvsplat, splatt3r, noposplat} that also use ground truth extrinsics for training and perform pose-free reconstruction at the inference stage. The additional pose supervision leverages the extrinsic information more effectively during training. \par

\subsubsection{Attribute Branch}

After obtaining geometry information from the geometry branch, we also need appearance and semantics information to complete the reconstructed 3D scenes and semantic field. Since appearance information, such as color and texture, is closely related to semantics, we handle them both in a single attribute branch but with two separate heads to better predict appearance and semantics distinctly:
\begin{equation}
\text{Attribute Branch} \left\{
\begin{aligned}
\text{Appearance Head}(D(f^v), img^v) & \rightarrow \{\Sigma_i^v, \alpha_i^v, c_i^v\}_{i=1}^{h \times w} \\
\text{Semantic Head}(D(f^v), img^v, F^v) & \rightarrow \{feat_i^v\}_{i=1}^{h \times w}
\end{aligned}
\right.
\quad
v \in \{0, 1\},
\end{equation}\par
where $img$ is the original image, $F$ is the semantic feature map of the original image, $\Sigma$, $\alpha$, and $c$ are the original components of the Gaussian that represent the appearance information, and $feat$ is the additional semantic feature that represents the semantic information of each Gaussian. \par

\textbf{Appearance Head}. The appearance head predicts original attributes for each Gaussian, including a 3D covariance matrix $\Sigma$, an opacity $\alpha_k$, and color $c$. These attributes, along with the position $\mu$ from the geometry branch, combine the original Gaussians and reconstruct 3D scenes. We render the novel view image from the reconstructed scenes and apply the photometric loss for training:
\begin{equation}\label{eq: photometric}
    \mathcal{L}_{photo} = \eta \frac{1 - SSIM(I, \hat{I})}{2} + (1 - \eta) \lVert I - \hat{I}\rVert_{2}, 
\end{equation}
where $I$ and $\hat{I}$ are the ground truth and 
rendered images, SSIM is the structural similarity index~\cite{ssim}. 

\textbf{Semantic Head}. To embed semantics into a 3D representation, the semantic head predicts an N-dimensional feature for each Gaussian. The appearance attributes from the appearance head and the semantic feature are combined into the semantic anisotropic Gaussians~\cite{feature3dgs}, which allows for rendering novel view feature maps. Thus, we unify the 3D scene and semantic field reconstruction. To supervise the novel view feature maps, we distill the feature from a pre-trained 2D semantic model~\cite{lseg} and apply a semantic distillation loss:
\begin{equation}
    \mathcal{L}_{sem} = 1 - \frac{\hat{feat} \cdot feat}{\lVert\hat{feat}\rVert \lVert feat \rVert},
\end{equation}
where $\hat{feat}$ is the predicted semantic feature and $feat$ is the feature from pre-trained 2D semantic model. By leveraging the pre-trained model, we eliminate the need for labeled 2D segmentation masks, which are high-cost to collect. Besides, since the pre-trained 2D semantic model is open vocabulary, we can apply the lightweight decoder to obtain novel view segmentation masks from reconstructed semantic fields with open text queries. \par

\textbf{Shortcut}. For both the appearance head and semantic head, we find that the original image or its semantic feature map preserves high-frequency details such as fine-grained color, texture, and edge information that are difficult to recover from the backbone feature $f$ alone. Therefore, we introduce a shortcut connection to facilitate the reconstruction of the high-frequency information. For the appearance head, we add the original image $img$ to the DPT feature $D(f)$. For the semantic head, we add both the original image $img$ and its semantic feature map $F$ to the DPT feature $D(f)$. This shortcut allows the model to leverage more original information to enhance its performance. \par

\subsection{Loss-guided View Sampler}\label{sampler}

Datasets for feed‑forward reconstruction are typically from video data. We select sparse frames as input context views and their corresponding rendered target views to compute the photometric loss $L_{photo}$ and the semantic distillation loss $L_{sem}$ during training. Existing methods~\cite{mvsplat, pixelsplat, splatt3r, noposplat} usually adopt a simple random selection for context and target views or enforce a scheduled max frame gap, but this often yields minimal overlap in early training that weakens supervision. Such minimal overlap can also cause Gaussians to expand incorrectly at view boundaries, degrading reconstruction quality. Some methods~\cite{lsm, splatt3r} use ground truth depth to obtain masks of the overlap regions to solve this problem. While relying on ground truth or masks restricts the training to specific datasets and limits the usage of large-scale video data. \par

To this end, we propose a loss-guided view sampler that avoids minimal view overlap during early training without relying on ground truth depth or masks. We constrain the angle between context views during early training and gradually increase it based on the loss. To alleviate the problem of scale ambiguity~\cite{pixelsplat} caused by pure image input without depth, we normalize the distance between input views and reconstruct the scaled scenes. This sample strategy avoids insufficient overlap views at the early stage of training and allows the model to adapt to larger view angles in later stages. The detailed algorithm flow is presented in Algorithm~\ref{algorithm}. \par

\begin{algorithm}
\small
\caption{Loss-guided View Sampler}
\label{algorithm}
\begin{algorithmic}[1]
\STATE \textbf{Input:} $Frame$, $L$, $i$, $\theta$, $\Delta \theta$
\STATE \textbf{Output:} $c_1$, $c_2$, $t$
\IF{$L$ is stable based on sliding window}
\STATE $\theta \gets \theta + \Delta \theta$
\STATE clear $L$
\ENDIF
\REPEAT
\STATE $max\_gap \gets Schedule(i)$
\STATE $gap \gets Random(2, max\_gap)$
\STATE $c_1 \gets Random(1, len(Frame) - gap)$
\STATE $c_2 \gets c_1 + gap$
\STATE $t \gets Random(c_1, c_2)$
\UNTIL $Angle(c_1, c_2) < \theta$ and $Angle(c_1, t) < \theta$ and $Angle(c_2, t) < \theta$
\STATE \textbf{return} $Frame[c_1], Frame[c_2], Frame[t]$
\end{algorithmic}
\end{algorithm}

For simplicity, we describe the selection of two context views $c_1$, $c_2$, and one target view $t$. Given a list of frames $Frame$ in the dataset, the sampler selects $c_1$, $c_2$, $t$ based on the current training step $i$ and the recent pose loss list $L$ based on a sliding window. We use the pose loss $L_{pose}$ from the geometry branch to guide the sampler since it can indicate the model has learned the correct relative pose of the context views and is ready for harder training cases. First, the maximum number of frames between context views is determined by a schedule based on the current training step $i$. Then, we select both context and target views with rotation angles below the dynamically adjusted threshold angle $\theta$, avoiding excessively large angular differences between views that cause insufficient overlap and ineffective supervision. The threshold angle $\theta$ is initialized before training and updated based on the recent pose loss list $L$ during training. When the geometry branch learns more accurate poses and the loss list $L$ stabilizes, the threshold angle increases by a small angle $\Delta \theta$, allowing the model to gradually adapt to more challenging context views. This progressive learning strategy enables the model to better capture geometric relationships between views, thereby improving the performance of pose-free reconstruction. \par

\subsection{Optimization}\label{optimization}

Through the Gaussian center positions from the geometry branch and other Gaussian attributes along with additional semantic features from the attribute branch, we unify the reconstruction of 3D scenes and semantic fields from sparse-view images. Novel view images and novel view feature maps are rendered via a differentiable splatting process. The whole model can be trained end-to-end with the following training objectives:
\begin{equation}
    \mathcal{L} = \mathcal{L}_{photo} + \lambda_{pose}\mathcal{L}_{pose} + \lambda_{sem}\mathcal{L}_{sem},
\end{equation}

where $\lambda_{pose}$ and $\lambda_{sem}$ are hyperparameters. $L_{photo}$ is the photometric loss for reconstruction. $L_{pose}$ serves as a proxy task to fully leverage the supervision signal of extrinsics and facilitate the learning of geometric. The $L_{sem}$ distills semantics from a pre‑trained 2D model to form a view‑consistent 3D semantic field, which enables rendering novel‑view feature maps and decoding them into dense segmentation masks with open text queries. \par

Upon completion of optimization, UniForward can reconstruct high-quality 3D scenes and semantic fields from only sparse-view images in about 0.1 seconds, and generalizes well across diverse scenes. \par

\section{Experiments}

\subsection{Implementation Details}

We use the ScanNet++ dataset~\cite{scannet++} with its official training and test splits, taking two input views at a resolution of $256 \times 256$. For a fair comparison, we retrained all the comparison methods under the same setting. For our method, we adopt the MASt3R backbone with its pre-trained weights while initializing all other parameters at random, and use the pre-trained 2D semantic model LSeg~\cite{lseg} for semantic distillation and novel view segmentation mask decoding, which is the same as our comparison methods~\cite{dff, feature3dgs, lsm} that also use LSeg. We train our model for 200,000 iterations using an Adam optimizer with a learning rate of $2 \times 10^{-5}$, and a batch size of 4. We set $\eta = 0.15$, $\lambda_{pose} = 0.1$, and $\lambda_{sem} = 0.1$ in the total loss function. To evaluate both 3D scene and semantic field reconstruction, we conduct novel view synthesis with peak signal-to-noise ratio (PSNR), structural similarity index (SSIM)~\cite{ssim}, and perceptual distance (LPIPS)~\cite{lpips} as metrics, and also conduct novel view segmentation in which scene labels are mapped to seven classes (wall, floor, ceiling, chair, table, door, and others) with mean intersection over union (mIoU) and mean pixel accuracy (mAcc) as metrics. All experiments are conducted on a single A6000 GPU with 48 GB. \par

\subsection{Comparison with Feed-Forward Methods}

We first compare our method against recent advanced feed-forward methods, including pixelSplat~\cite{pixelsplat}, MVSplat~\cite{mvsplat}, LSM~\cite{lsm}, Splatt3R~\cite{splatt3r}, and NoPoSplat~\cite{noposplat}, along with the pre-trained 2D semantic model LSeg~\cite{lseg} that is used during training. The results are reported in Table~\ref{tab: main}. Since not all methods reconstruct both 3D scenes and semantic fields, we mark ``N/A'' where they cannot perform novel view synthesis or segmentation. Our UniForward outperforms other methods on both novel view synthesis and segmentation with fewer constraints. We also observe that our method outperforms the pre-trained 2D semantic model on novel view segmentation, probably because our method lifts semantics to 3D, ensuring spatial consistency, while the 2D model lacks this capability.\par

\begin{table}[!t]
\centering
\small
\caption{Comparison of our method against other feed-forward methods. ``N/A'' indicates that the method lacks the capability to perform the corresponding task.}
\begin{tabular}{l|ccc|ccc|cc}
    \toprule
    \multirow{2}{*}{\textbf{Method}} & \multirow{2}{*}{Pose-Free} & \multirow{2}{*}{\makecell{Depth-Free \\ (For Training)}} & \multirow{2}{*}{Intr.-Free} & \multicolumn{3}{c|}{Novel View Synthesis} & \multicolumn{2}{c}{Novel View Seg.} \\
    & & & & \textbf{PSNR$\uparrow$} & \textbf{SSIM$\uparrow$} & \textbf{LPIPS$\downarrow$} & \textbf{mAcc$\uparrow$} & \textbf{mIoU$\uparrow$} \\
    \midrule
    \textcolor{gray}{LSeg} & \textcolor{gray}{N/A} & \textcolor{gray}{N/A} & \textcolor{gray}{N/A} & \textcolor{gray}{N/A} & \textcolor{gray}{N/A} & \textcolor{gray}{N/A} & \textcolor{gray}{\underline{0.739}} & \textcolor{gray}{\underline{0.340}} \\
    pixelSplat & \ding{55} & \ding{51} & \ding{55} & 21.747 & 0.786 & 0.225 & N/A & N/A \\
    MVSplat & \ding{55} & \ding{51} & \ding{55} & 22.649 & 0.775 & 0.203 & N/A & N/A \\
    LSM & \ding{51} & \ding{55} & \ding{51} & 16.005 & 0.630 & 0.388 & 0.674 & 0.320 \\
    Splatt3R & \ding{51} & \ding{55} & \ding{51} & 19.050 & 0.659 & 0.275 & N/A & N/A \\
    NoPoSplat & \ding{51} & \ding{51} & \ding{55} & \underline{24.755} & \underline{0.814} & \underline{0.190} & N/A & N/A \\
    \textbf{Ours} & \ding{51} & \ding{51} & \ding{51} & \textbf{26.147} & \textbf{0.850} & \textbf{0.149} & \textbf{0.740} & \textbf{0.347} \\
    \bottomrule
\end{tabular}
\label{tab: main}
\end{table}

\begin{table}[!t]
\centering
\small
\caption{Comparison of our method against per-scene optimized methods.}
\begin{tabular}{l|c|c|ccc|cc}
    \toprule
    \multirow{2}{*}{\textbf{Method}} & \multirow{2}{*}{Feed-Forward} & \multirow{2}{*}{\makecell{Reconstrucion \\ Time}} & \multicolumn{3}{c|}{Novel View Synthesis} & \multicolumn{2}{c}{Novel View Seg.} \\
    & & & \textbf{PSNR$\uparrow$} & \textbf{SSIM$\uparrow$} & \textbf{LPIPS$\downarrow$} & \textbf{mAcc$\uparrow$} & \textbf{mIoU$\uparrow$} \\
    \midrule
    Feature 3DGS & \ding{55} & $\approx$ 10min & 13.966 & 0.436 & 0.537 & 0.605 & 0.250 \\
    DFFs & \ding{55} & \underline{$\approx$ 2min} & \underline{18.140} & \underline{0.592} & \underline{0.436} & \underline{0.697} & \underline{0.287} \\
    \textbf{Ours} & \ding{51} & \textbf{0.105s} & \textbf{28.097} & \textbf{0.881} & \textbf{0.103} & \textbf{0.758} & \textbf{0.345} \\
    \bottomrule
\end{tabular}
\label{tab: scene-optimized}
\end{table}

\subsection{Comparison with Per-Scene Optimized Methods}

We further compare our method with per-scene optimized methods, including NeRF-based DFFs~\cite{dff} and 3DGS-based Feature 3DGS~\cite{feature3dgs}. These per-scene optimized methods rely on Structure-from-Motion (SfM)~\cite{sfm} to obtain camera extrinsic and intrinsic parameters and are not designed for sparse-view input. Since SfM may fail on images with insufficient overlap, we select scenes from the test dataset that can be successfully reconstructed by these comparison methods. As shown in Table~\ref{tab: scene-optimized}, our method outperforms these per-scene optimization methods in both novel view synthesis and segmentation. Additionally, per-scene optimized methods take several minutes to optimize a single scene. In contrast, our method reconstructs both the 3D scene and its semantic field in only about 0.1 seconds, and achieves significantly better performance.\par

\begin{figure}[!t]
    \centering
    \includegraphics[width=0.95\linewidth]{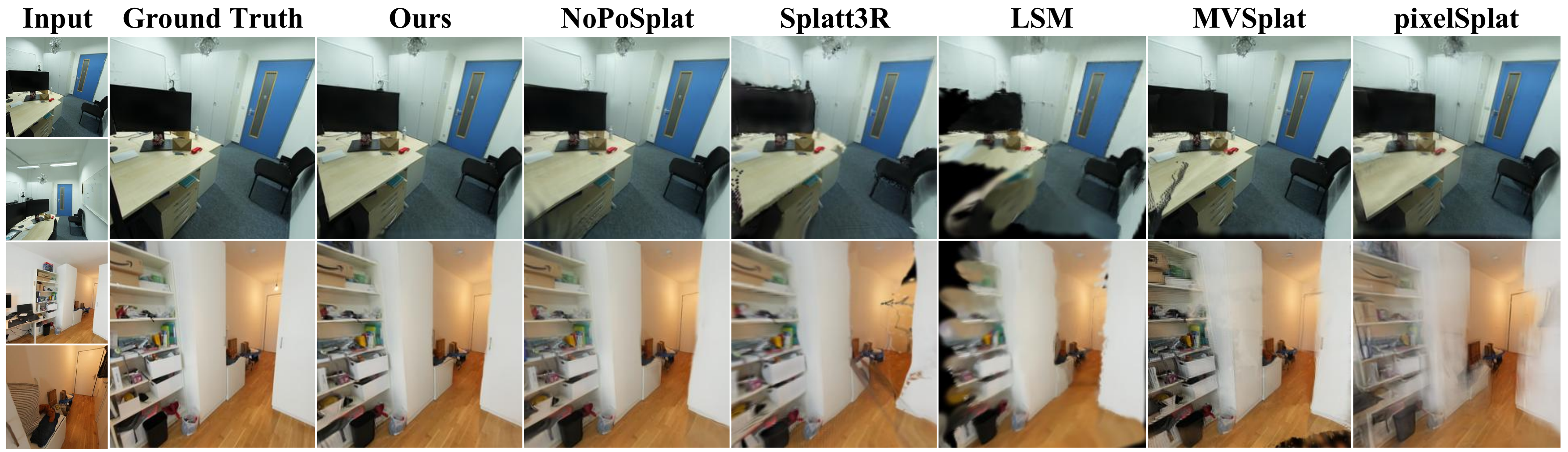}
    \caption{Qualitative comparison on novel view synthesis.}
    \label{fig: syn}
\end{figure}

\begin{figure}[!t]
    \centering
    \includegraphics[width=0.95\linewidth]{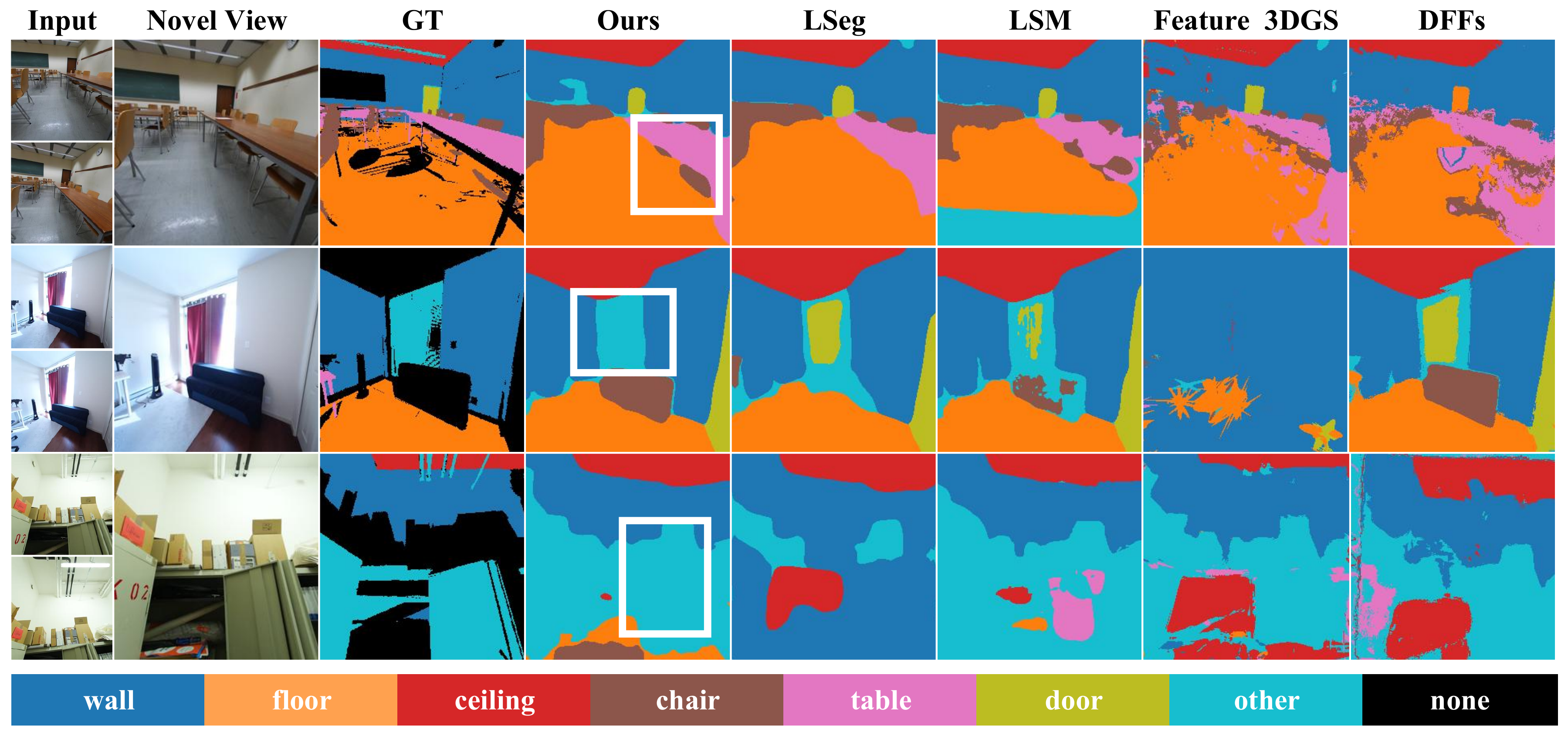}
    \caption{Qualitative comparison on novel view segmentation.}
    \label{fig: seg}
\end{figure}

\subsection{Qualitative Comparison}
Figure~\ref{fig: syn} and Figure~\ref{fig: seg} present qualitative comparisons on novel view synthesis and segmentation. For novel view synthesis, our method achieves superior geometry reconstruction (e.g., the wall in the 2nd row) and can handle both large (1st row) and small (2nd row) overlap input. For novel view segmentation, our method outperforms both feed-forward and per-scene optimized methods. Furthermore, our method outperforms the pre-trained 2D semantic model~\cite{lseg} used for training, demonstrating improved 3D consistency through unified 3D scene and semantic field reconstruction. \par

\subsection{Ablation Studies}

\begin{table}[t!]
\caption{Ablation studies. All ablation models are based on our full model by removing the corresponding module.}
\centering
\begin{tabular}{rl|ccc|cc}
    \toprule
    & \multirow{2}{*}{Variant Model} & \multicolumn{3}{c|}{Novel View Synthesis} & \multicolumn{2}{c}{Novel View Seg.} \\
    & & PSNR$\uparrow$ & SSIM$\uparrow$ & LPIPS$\downarrow$ & mAcc$\uparrow$ & mIoU$\uparrow$ \\
    \midrule
    1 & Full Model & 26.147 & 0.850 & 0.149 & 0.740 & 0.347 \\
    2 & w/o Loss-guided Sampler & 25.690 & 0.841 & 0.161 & 0.729 & 0.341 \\
    3 & w/o Pose Head & 25.900 & 0.848 & 0.151 & 0.732 & 0.342  \\
    4 & w/o Sem. Head & 25.455 & 0.842 & 0.154 & 0.652 & 0.317 \\
    5 & w/o Shortcut & 24.321 & 0.807 & 0.232 & 0.698 & 0.325 \\
    \bottomrule
\end{tabular}
\label{tab: ablation}
\end{table}

We conduct ablation studies to comprehensively analyze the effectiveness of our method, as shown in Table~\ref{tab: ablation}. The ablation variants are derived by removing specific modules from the original full model (1st row). We replace our loss-guided view sampler with a simple random selection that has a scheduled max gap (2nd row). The results demonstrate that our loss-guided view sampler effectively stabilizes training and improves performance. Besides, we validate the effectiveness of the dual-branch decoupled decoder. By predicting camera poses as a proxy task, the pose head enables the model to better understand 3D scene geometry and enhance the performance of novel view synthesis (3rd row). The semantic head also improves performance, particularly in semantic field reconstruction, and gains notable improvements on novel view segmentation (4th row). The shortcut in the attribute branch facilitates both 3D scene and semantic field reconstruction (5th row), demonstrating the importance of additional information provided by raw images and features. \par

\section{Conclusion, Limitations, and Broader Impacts}
In this paper, we introduce UniForward, a feed-forward Gaussian Splatting model that unifies 3D scene and semantic field reconstruction. For each view, we propose a dual-branch decoupled decoder to predict Gaussian components and propose a loss-guided view sampler for training view selection. Our model requires no ground truth depth during training. Semantics are learned from pre-trained 2D semantic models without semantic labels, and other Gaussian attributes are supervised by photometric loss. At the inference stage, UniForward achieves real-time 3D scene and semantic field reconstruction from sparse-view images and generalizes across diverse scenes. \par

Although our model successfully unifies the reconstruction of 3D scenes and semantic fields, it only reconstructs within input‐view coverage and cannot reconstruct the unseen regions, and the pre‑trained 2D semantic model constrains the fidelity of the reconstructed semantic field. In addition, unifying the 3D scene and semantic field increases model parameters and computational overhead. \par

This work enables more accurate environment understanding for applications such as robotic navigation and AR/VR. It also provides high‑quality, on‑the‑fly 3D models with semantic labels for digital‑twin and industrial simulation scenarios. However, realistic reconstructions may be misused for deepfakes or fraud. We recommend enforcing access controls, maintaining usage logs, and establishing clear licensing to prevent misuse. \par

\bibliography{main}
\bibliographystyle{abbrvnat}

\newpage
\clearpage

\begin{center}
\LARGE\bf
    Supplementary Material\par
\vspace{20pt}
\end{center}
\renewcommand\thesection{\Alph{section}}
\setcounter{section}{0}
\setcounter{figure}{0}

\section{Additional Implementation Details}

\subsection{Datasets}

\textbf{ScanNet++}~\cite{scannet++} is a large-scale dataset of indoor scenes containing sub-millimeter resolution laser scans. The reconstructed scenes are annotated with semantic labels, providing high-quality 3D scenes with corresponding semantic fields. We use the official training set and use the validation set for testing since the official test set has no semantic annotation. We exclude the scenes that do not have enough frames. Our training set contains 230 scenes, and our test set contains 49 scenes. \par 

\textbf{ScanNet}~\cite{scannet} is also a large-scale dataset of indoor scenes with 3D semantic annotation. Since it contains different scenes and uses different collection equipment from that of ScanNet++, we conduct additional experiments on ScanNet to demonstrate the broad applicability of our method. We follow the official data splits in ScanNet. Since some scenes contain invalid poses, we select 200 scenes from the validation dataset as our test set to exclude these invalid scenes. \par

\subsection{Metrics}

To evaluate the quality of the reconstructed scenes and semantic fields, we perform the task of novel view synthesis and novel view segmentation, respectively. For novel view synthesis, we follow the general peak signal-to-noise ratio (PSNR), structural similarity index (SSIM)~\cite{ssim}, and perceptual distance (LPIPS)~\cite{lpips} as metrics. For novel view segmentation, since we reconstruct the 3D semantic field, we first render the 2D novel view feature map from the novel viewpoint. We then select a set of common categories in the dataset, including ``wall'', ``floor'', ``ceiling'', ``chair'', ``table'', ``door'', and ``other''. Using these category names as text inputs, we employ the decoder of the pre-trained 2D open-vocabulary semantic model, LSeg~\cite{lseg}, to obtain the novel view segmentation mask. We evaluate the novel view segmentation mask against the ground truth using standard segmentation metrics, including mean Intersection-over-Union (mIoU) and mean pixel Accuracy (mAcc). The evaluation for reconstructed semantic fields is the same as our comparison methods~\cite{lsm, dff, feature3dgs}, ensuring fair comparisons. \par


\subsection{Comparison Methods}

We compare our method with both feed-forward methods and per-scene optimized methods. The feed-forward methods includes pixelSplat~\cite{pixelsplat}, MVSplat~\cite{mvsplat}, LSM~\cite{lsm}, Splatt3R~\cite{splatt3r}, and NoPoSplat~\cite{noposplat}. For LSM, we directly use the official pre-trained weights, as the model is trained on ScanNet and ScanNet++, which include our datasets. For other methods, we retrain them on the ScanNet++ dataset with their original configurations. The per-scene optimized methods include DFFs~\cite{dff} and Feature 3DGS~\cite{feature3dgs}. We optimize these methods scene by scene on our test set with their original configurations. \par

\begin{figure}[!t]
    \centering
    \includegraphics[width=0.9\linewidth]{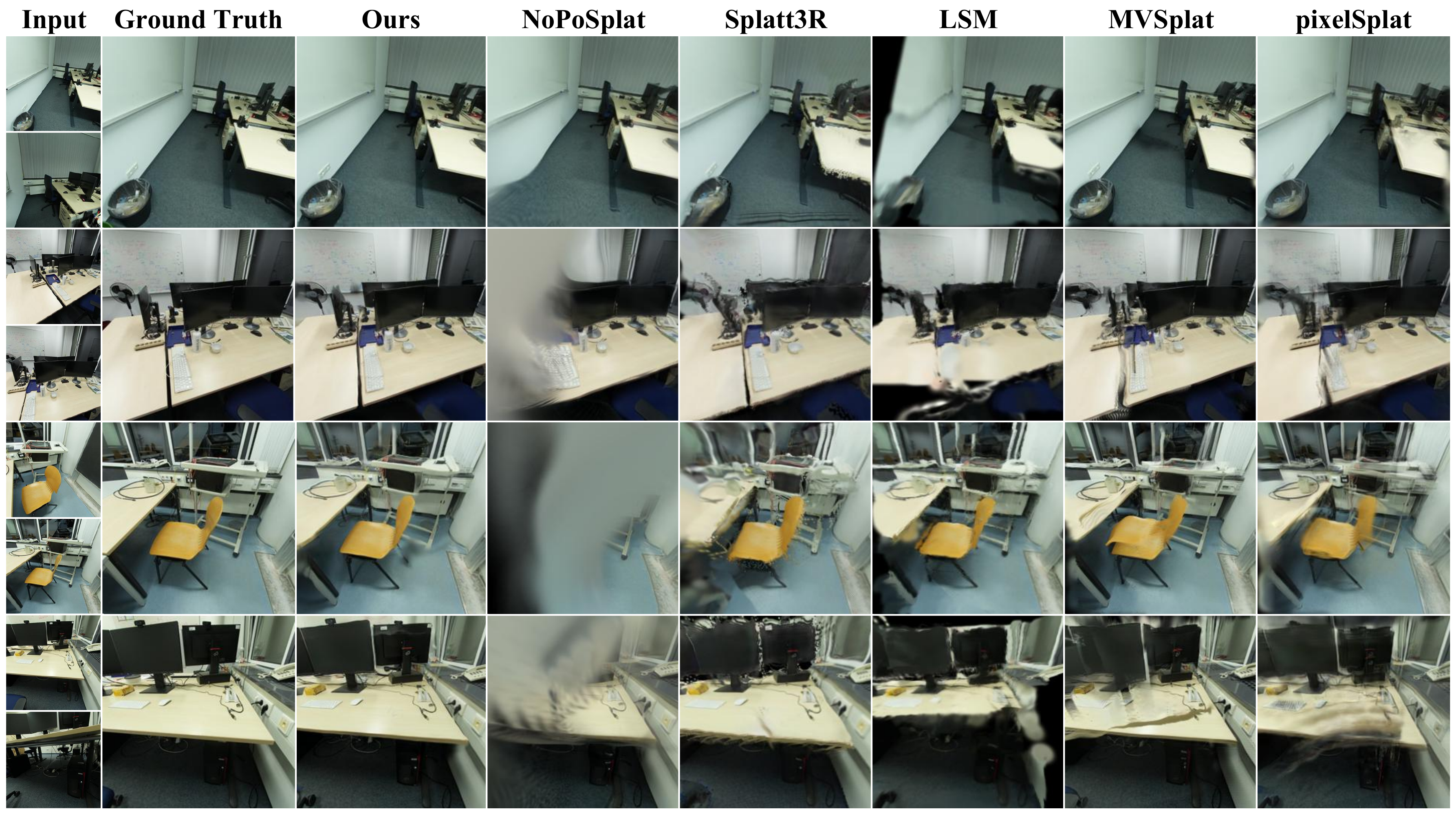}
    \caption{Additional qualitative comparison results on novel view synthesis. Our method avoids the incorrect expansion of Gaussians at view edges, while comparison methods such as NoPoSplat do not. Therefore, we synthesize higher-quality novel view images, indicating better reconstruction of 3D scenes.}
    \label{fig: syn_supple}
\end{figure}

\begin{figure}[!t]
    \centering
    \includegraphics[width=0.9\linewidth]{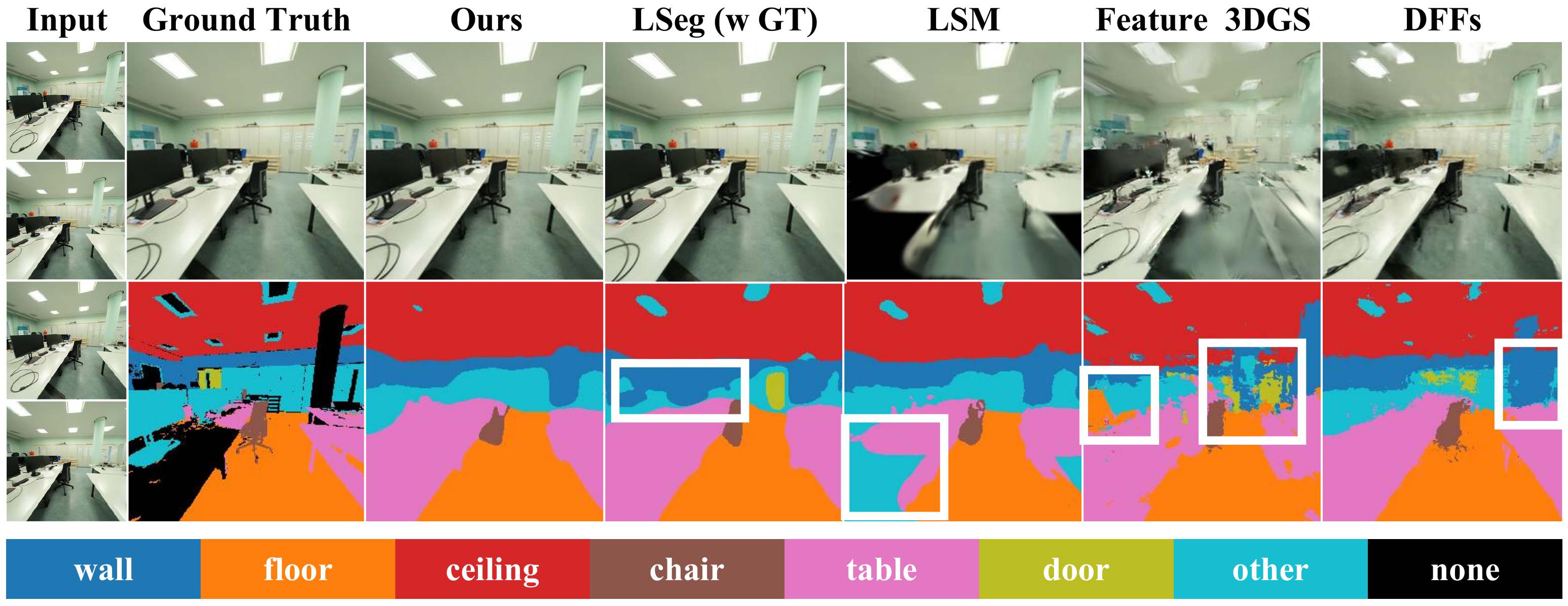}
    \caption{Additional qualitative comparison results on novel view synthesis and segmentation. The 3D semantic field reconstructed by our method enables the novel view segmentation masks with smoother edges. In contrast, other methods often suffer from partial omissions or jagged edges due to lower reconstruction quality. Besides, our method outperforms the 2D semantic model LSeg, even though LSeg uses GT views as input. This is probably because our reconstructed 3D semantic field achieves spatial consistency, while the 2D model lacks this capability.}
    \label{fig: seg_supple}
\end{figure}

\section{Additional Qualitative Results on ScanNet++}

We provide additional qualitative comparison results of novel view synthesis and novel view segmentation on ScanNet++ in Figure~\ref{fig: syn_supple} and Figure~\ref{fig: seg_supple}. \par

We also provide \textbf{video demos} in the supplementary material, which contain the visual results of the reconstructed 3D scenes, semantic fields, and geometry fields on ScanNet++. The video demos involve the comparison of our method against LSM~\cite{lsm}. LSM recovers scene scale through ground truth depth. However, recovering scene scale from depth can lead to a mismatch between the camera frustum and the reconstructed scene, resulting in situations where the camera penetrates the scene boundaries. Moreover, the reconstruction quality of LSM is suboptimal, lacking 3D consistency. In contrast, our method does not suffer from these issues. \par

\begin{table}[!t]
\centering
\small
\caption{Additional comparison on ScanNet dataset. Our UniForward also outperforms other methods on the other dataset, demonstrating the broad applicability of our method.}
\begin{tabular}{l|ccc|ccc|cc}
    \toprule
    \multirow{2}{*}{\textbf{Method}} & \multicolumn{3}{c|}{Novel View Synthesis} & \multicolumn{2}{c}{Novel View Seg.} \\
    & \textbf{PSNR$\uparrow$} & \textbf{SSIM$\uparrow$} & \textbf{LPIPS$\downarrow$} & \textbf{mAcc$\uparrow$} & \textbf{mIoU$\uparrow$} \\
    \midrule
    \textcolor{gray}{LSeg} & \textcolor{gray}{N/A} & \textcolor{gray}{N/A} & \textcolor{gray}{N/A} & \textcolor{gray}{0.762} & \textcolor{gray}{0.331} \\
    LSM & 19.083 & 0.698 & 0.358 & 0.762 & 0.341 \\
    \textbf{Ours} & \textbf{23.361} & \textbf{0.764} & \textbf{0.238} & \textbf{0.783} & \textbf{0.356} \\
    \bottomrule
\end{tabular}
\label{tab: scannet}
\end{table}

\section{Additional Results on ScanNet}

To demonstrate the broad applicability of our UniForward, we conduct additional experiments on the ScanNet dataset and compare our method with LSM~\cite{lsm}. As presented in Table~\ref{tab: scannet}, our method outperforms the comparison method LSM on both novel view synthesis and novel view segmentation, and also achieves superior performance than the pre-trained 2D semantic model LSeg used for training. \par

\section{Additional Qualitative Results on ScanNet.}

\begin{figure}[!t]
    \centering
    \includegraphics[width=0.8\linewidth]{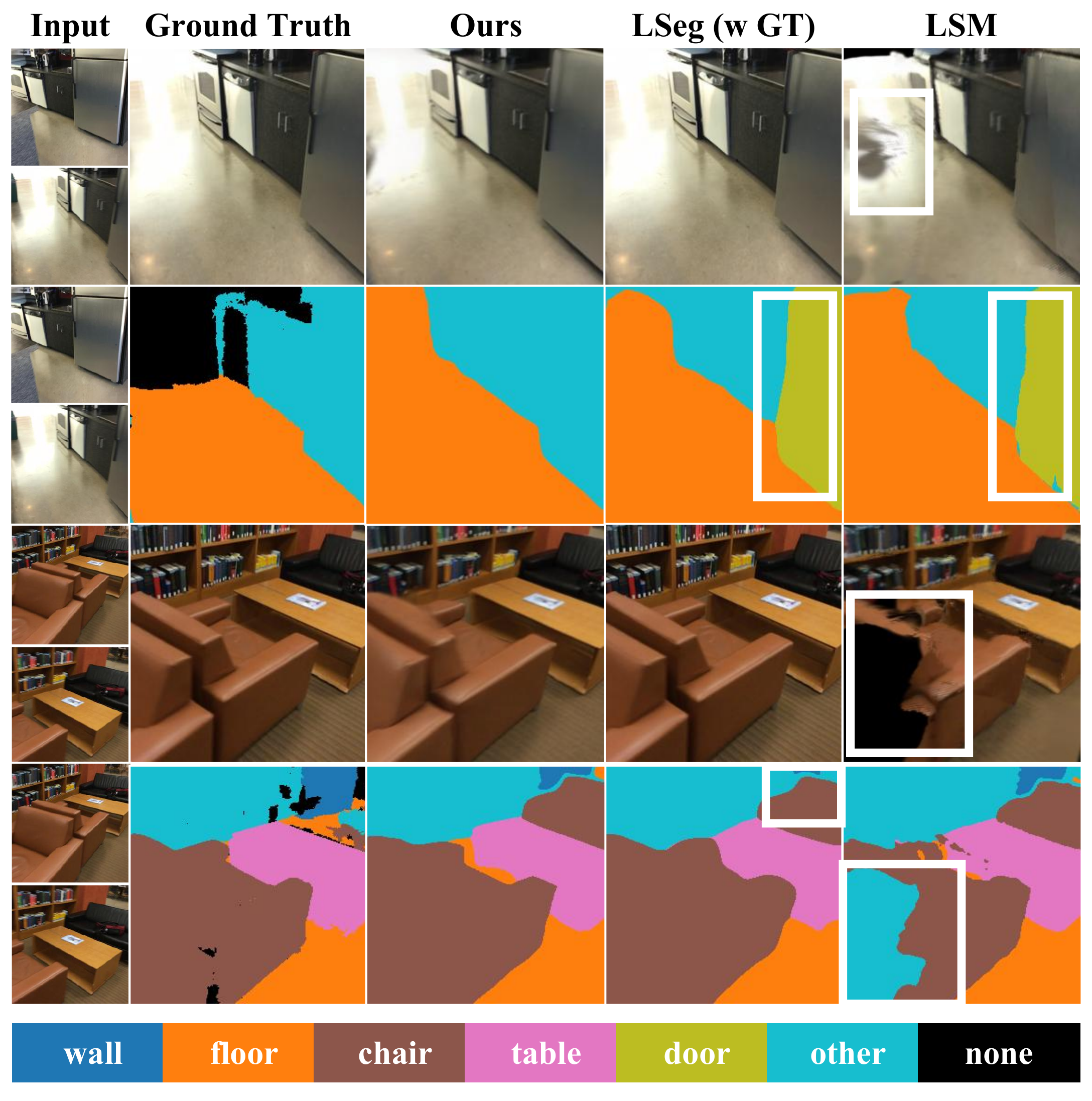}
    \caption{Additional qualitative comparison on ScanNet. Our method produces more accurate semantic embeddings while preserving spatial consistency, thereby outperforming both the comparison method, LSM, and the LSeg model used during training.}
    \label{fig: scannet}
\end{figure}

We provide qualitative comparison results of novel view synthesis and novel view segmentation on ScanNet in Figure~\ref{fig: scannet}. Our method achieves better visual results than LSM on both tasks and outperforms the 2D semantic model LSeg on novel view segmentation,  demonstrating the quality of our reconstructed 3D scenes and semantic fields. \par

\end{document}